\documentclass[11pt]{article}
\usepackage[usenames,dvipsnames,svgnames,table]{xcolor} 

\usepackage{algorithm}
\usepackage{algorithmic}
\floatname{algorithm}{Algorithm}

\usepackage{longtable}
\usepackage{hhline}
\usepackage{float}
\newfloat{algorithm}{tb}{lop}

\usepackage{enumitem} 
\usepackage{rotfloat} 
\usepackage{graphicx}
\usepackage{epsfig}
\usepackage{amssymb, amsmath, amsthm}
\usepackage{verbatim}
\usepackage{natbib}
\usepackage{authblk}
\usepackage{multirow}
\usepackage[colorlinks]{hyperref} 
\usepackage[colorinlistoftodos, textsize=scriptsize]{todonotes} 
\usepackage{subcaption} 
\usepackage{array}
\newcolumntype{L}{>{\centering\arraybackslash}m{0.1\linewidth}}

\newtheorem{theorem}{Theorem}[section]
\newtheorem{lemma}[theorem]{Lemma}
\newtheorem{proposition}[theorem]{Proposition}

\newtheorem{definition}{Definition}

\newcommand{\bbR}{\mathbb{R}}

\newcommand{\bbN}{\mathbb{N}}
\newcommand{\bbX}{\mathbb{X}}

\newcommand{\bfL}{\mathbf{L}}

\newcommand{\norm}[1]{\left\lVert#1\right\rVert}

\newcommand{\N}{\mathbb{N}} 
\newcommand{\R}{\mathbb{R}} 
\newcommand{\im}{\operatorname{im}}

\newcommand{\matL}{\mathbf{L}} 
\DeclareMathOperator*{\diam}{diam} 
\newcommand{\httt}{\operatorname{height}} 

\newcommand{\ptcloudx}{\mathbb{X}} 
\newcommand{\ptcloudy}{\mathbb{Y}} 

\newcommand{\veca}{\mathbf{a}} 
\newcommand{\vecb}{\mathbf{b}} 
\newcommand{\vecv}{\mathbf{v}} 

\begin{document}
	
\title{Shape-Preserving Dimensionality Reduction : An Algorithm and Measures of Topological Equivalence}
\author[1]{Byeongsu Yu}
\affil[1]{Dept. of Mathematics, Texas A\&M University}
\author[2]{Kisung You}
\affil[2]{Dept. of ACMS, University of Notre Dame}

\date{}

\maketitle
\begin{abstract}
We introduce a linear dimensionality reduction technique preserving topological features via persistent homology. The method is designed to find linear projection $L$ which preserves the persistent diagram of a point cloud $\ptcloudx$ via simulated annealing. The projection $L$ induces a set of canonical simplicial maps from the Rips (or \v{C}ech) filtration of $\ptcloudx$ to that of $L\ptcloudx$. In addition to the distance between persistent diagrams, the projection induces a map between filtrations, called filtration homomorphism. Using the filtration homomorphism, one can measure the difference between shapes of two filtrations directly comparing simplicial complexes with respect to quasi-isomorphism $\mu_{\operatorname{quasi-iso}}$ or  strong homotopy equivalence $\mu_{\operatorname{equiv}}$. These $\mu_{\operatorname{quasi-iso}}$ and $\mu_{\operatorname{equiv}}$ measures how much portion of corresponding simplicial complexes is quasi-isomorphic or homotopy equivalence respectively. We validate the effectiveness of our framework with simple examples.
\end{abstract}

\section{Introduction}\label{sec:intro}

Dimensionality reduction is one of the most important pillars in statistics and machine learning that aims to reveal patterns embedded in the data and has attained long-dated interests with a vast literature \citep{engel_survey_2012, ma_review_2013-2}. The purpose of dimensionality reduction is not limited to visualization of the high-dimensional data, but it plays a crucial role as a preliminary step for further analysis \citep{jolliffe_note_1982-1}. For example, independent component analysis (ICA) is a decomposition algorithm for a multivariate signal into a sum of statistically independent signals \citep{hyvarinen_independent_2001-1, hyvarinen_independent_2013}. In the field of brain connectomics, ICA is a gold standard in extracting spatial activation patterns in high-dimensional observations of functional magnetic resonance imaging data and enables tractable analysis with low-dimensional embedding  \citep{mckeown_independent_2003-1, park_structural_2013-2}. 

Since the advent of principal component analysis (PCA) by \cite{pearson_liii._1901}, a large number of algorithms have been proposed that are based on a variety of principles on how and what aspects of information should be addressed from the data \citep{you_rdimtools_2020}. Many popular algorithms nowadays are based on a so-called manifold assumption that the data lies on a low-dimensional manifold in a high-dimensional space. In layman's terms, this class of methods tries to capture the shape of data. For example, Laplacian eigenmaps approximates the data manifold via eigenfunctions of a discrete analog for the Laplace-Beltrami operator on a neighborhood graph that is reconstructed on a point cloud in the high-dimensional space \citep{belkin_laplacian_2001}. Isomap reveals the low-dimensional structure of the data by applying multidimensional scaling onto the metric graph via the shortest path on a neighborhood graph of the data \citep{tenenbaum_global_2000}, while locally linear embedding considers a sparse connectivity graph obtained by reconstruction of each observation by its neighbors \citep{roweis_nonlinear_2000}. These algorithms share a common idea to discover geometric structures underlying the data and have influenced many to deliver numerous algorithms thereafter. 

Recently, a line of study called topological data analysis (TDA) has gained much attention that refers to a class of methods to study the shape of data from perspectives of topology \citep{MR2549932, MR2572029, MR3328629}. Persistent homology is a central tool of TDA to quantify the topological properties of the data in a multiscale manner. Some recent works have used persistent homology as a quality measure of an embedding \citep{drevaluation, MR3822886, Bastian17}, which have fundamental limitation if one wants to use persistent homology as an objective itself. The work of \cite{PAUL2017160} is another use case of persistent homology as a benchmark measure to evaluate the efficacy of dimensionality reduction algorithms. Moreover, \cite{sheehy14persistent} showed that a certain random projection preserving critical points also preserves the persistent homology. Along with this result, \cite{aryaetal} shows that every linear projection map preserves the \v{C}ech complex arising from a specially weighted distance between a point and its point cloud. 

We propose a framework to find a linear map for dimension reduction that preserves topological features of data by choosing a linear projection minimizing the Wasserstein distance between persistent diagrams of the original and projected data. Moreover, the choice of linear projection allows us to compare not only persistent diagrams but also simplicial complexes directly via a \emph{filtration homomorphism}, which is a specific set of simplicial maps between filtrations. This homomorphism enables us to investigate whether each simplicial map induces quasi-isomorphism or strong equivalence by Theorem \ref{classification_thm}, whose information is summarized as similarity measures $\mu_{\operatorname{quasi-iso}}$ and $\mu_{\operatorname{equiv}}$ that account for how much portion of intervals the filtration homomorphism induces quasi-isomorphism or homotopy equivalence. 

Our approach is different from that of \cite{aryaetal} whose goal is to find an ``optimized'' distance-preserving \v{C}ech complex working for all linear projections. On the other hand, we center on finding an ``optimized'' linear projection that preserves the \v{C}ech complex under the Euclidean distance. Also, we recently came across the work by \cite{kachan_persistent_2020} that utilizes the same objective function. However, we provide a gradient-free optimization routine to reduce computational overhead and propose quality measures on top of principled approaches to quantify topological differences.

The rest of the paper is organized as follows. Section \ref{sec:prelim} delivers a brief introduction to mathematical backgrounds for TDA. Section \ref{sec:algorithm} presents the \textsc{spred} algorithm that aims to find a linear projection preserving persistent homology. In Section \ref{sec:theory}, we propose a systematic framework to measure the difference between two filtrations using topological notions of quasi-isomorphism, weak and strong homotopy equivalences. Examples with simulated and real data are given in Section \ref{sec:example} to evaluate the efficacy of our framework. We conclude in Section \ref{sec:conclusion} by highlighting the unique advantages of our framework and discuss potential directions of extension for future studies.

\section{Preliminaries}\label{sec:prelim}

We begin this section by introducing notations. $\bbN$ denotes a set of nonnegative integers $\{ 0,1,2,\cdots\}$. Lower case letters $n,m,\cdots$ denote elements in $\bbN$ while letters such as $a,b, \cdots$ denote elements in $\bbR_{\geq0}$, a set of all non-negative real numbers. An upper case letters $A,B, \cdots$ denotes a set. $\diam A$ is a \emph{diameter} of a set $A$, defined as the largest pairwise distance among all pairs of $A$. $2^{A}$ denotes a power set of $A$. $\bbR^{n}$ is the $n$-dimensional real vector space. Boldface upper case letters and boldface lower case letters denote matrices and vectors, respectively. $|A|$ denotes a cardinality of $A$. Map denotes a function $f: A \to B$. The kernel of a function is denoted as $\ker f$, and the image of $f$ is denoted as $\im f$. A \emph{point cloud} $\ptcloudx$ is a finite set of vectors $\{x_{1},\cdots, x_{m} \} \subseteq \bbR^{n}$ for some $m\in \bbN.$ We assume that $\ptcloudx$ is a set of observations about $M$. Let $d_{\ptcloudx}$ be a \emph{distance function over $\ptcloudx$} defined as $d_{\ptcloudx}(a) = \inf_{x \in \ptcloudx}\norm{x-a}$, where $\norm{x}$ is the usual Euclidean norm. Lastly, $\mathbb{F}$ denotes a field.

Next, we provide a brief review of notions in TDA. We refer interested readers for an in-depth overview of the topic to standard references \citep{edelsbrunner,MR2549932,MR2572029,MR3328629}. One of the primary goals for TDA is to recover topological information of the underlying set of the point cloud by a series of abstract simplicial complexes. An (abstract) \emph{simplicial complex} $\mathcal{K}$ over $\ptcloudx$ be a subset of $2^{\ptcloudx}$ such that for any $\sigma \in \mathcal{K}$ and a subset $\tau$ of $\sigma$, $\tau \in \mathcal{K}$. Each element of $\mathcal{K}$ is called the \emph{face} of $\mathcal{K}$. Given two simplicial complexes $\mathcal{K}$ and $\mathcal{Q}$, a \emph{simplicial map} $\phi$ is a map sending vertices of $\mathcal{K}$ to those of $\mathcal{Q}$ such that if $\{ a_{1},\cdots, a_{m} \} \subset \ptcloudx$ forms a simplex in $\mathcal{K}$ then $\{ \phi(a_{1}),\cdots, \phi(a_{m}) \}$ is a simplex of $\mathcal{Q}$. The image of a simplex is nonexpanding since it may have a smaller cardinality than its preimage. 

There are two popular choices to build a simplicial complex of $\ptcloudx$ using the level set $d_{\ptcloudx}^{-1}(0,t]$ for some $t \in \bbR_{\geq0}$. A \emph{Rips} complex $R(t)$ is a set of all subsets $\sigma =\{ \veca_{1}, \veca_{2}, \cdots, \veca_{k} \} \in 2^{\ptcloudx}$ such that $\sigma$ lies in the same connected component of $d_{\ptcloudx}^{-1}(0,t]$. It can be defined equivalently by is $R(t):= \{ \sigma = \{ \veca_{1}, \veca_{2}, \cdots, \veca_{k}\} : \norm{\veca_{i}-\veca_{j}} \leq 2t \text{ for } \forall  i,j \in \{ 1,2, \cdots, k\} \}.$ On the other hand, a \emph{\v{C}ech} complex is a set of all subsets $\sigma =\{ \veca_{1}, \veca_{2}, \cdots, \veca_{k} \} \in 2^{\ptcloudx}$ such that $\sigma$ lies in the same connected component of $d_{\ptcloudx}^{-1}(0,t]$ and the component is contractible. An equivalent definition is $C(t):=\{ \sigma = \{ \veca_{1}, \veca_{2}, \cdots, \veca_{k}\} : \bigcap_{i=1}^{k}B_{t}(\veca_{i}) \neq\emptyset\}$ where $B_{t}(\veca_{i})$ is a ball of radius $t$ centered at $\veca_{i}$ in $\R^{n}$. For any simplicial complex, let $\mathcal{F}_{\ptcloudx}= \{ R(t) \text{ or }C(t) \}_{t \geq 0}$ be \emph{filtration of simplicial complexes}. Since both $R(t)$ and $C(t)$ are subcomplex of $2^{\ptcloudx}$ as a simplicial complex, $\mathcal{F}_{\ptcloudx}$ has finitely many distinct simplicial complexes. Thus, we may let $\mathcal{F}_{\ptcloudx}:= \{ \mathcal{K}_{i} \}_{i \in I}$ where $I$ is a finite cover of $\R_{\geq0}$ whose elements are disjoint half-open sets of the form $[a,b)$ for some $a<b \in \R_{\geq0}$. Thus, we may regard $I$ as a finite totally ordered set, and $\mathcal{K}_{[a,b)} \subseteq \mathcal{K}_{[a',b')}$ if $a<a'$. The nerve lemma \citep{munch} assures that each $\mathcal{K}_{i}$ is homeomorphic to $d_{\ptcloudx}^{-1}(0,t_{i}]$. 

One may summarize topological features of simplicial complexes in $\mathcal{F}_{\ptcloudx}=\{ \mathcal{K}_{[a,b)} \}_{[a,b) \in I}$ using (simplicial) $i$-th \emph{homology} groups $\{H_{i}(\mathcal{K}_{[a,b)};\mathbb{F})\}_{i=0}^{\infty}$ for each $[a,b) \in I$ with coefficients over $\mathbb{F}$. For any $i \in \N$, the $i$-th \emph{persistent homology} $PH_{i}(\ptcloudx ; \mathbb{F})$ is a set $\{H_{i}(\mathcal{K}_{[a,b)};\mathbb{F})\}_{[a,b) \in I}$ of homology groups over a filtration of the simplicial complexes. Given $[a,b) \in I$, a group homomorphism $\varphi_{b}:H_{i}(\mathcal{K}_{[a,b)};\mathbb{F}) \to H_{i}(\mathcal{K}_{[b,c)};\mathbb{F})$ is induced by a natural inclusion $\mathcal{K}_{[a,b)} \to \mathcal{K}_{[b,c)}$ for $[b,c) \in I$, the successor of $[a,b)$ in $I$. An element $\gamma \in H_{i}(\mathcal{K}_{[a,b)};\mathbb{F})$ is called an $i$-th \emph{homology class} if $\gamma$ is in a fixed basis of $H_{i}(\mathcal{K}_{[a,b)};\mathbb{F})$. Then, we say $\gamma$ is \emph{born} at $\mathcal{K}_{[a,b)}$ if $\gamma \not\in \im\varphi_{[z,a)}$ for the predecessor $[z,a) \in I$. Similarly, $\gamma$ is said to \emph{die} at $\mathcal{K}_{[c,d)}$ if the image of $\gamma$ from $H_{i}(\mathcal{K}_{[a,b)};\mathbb{F}) \to H_{i}(\mathcal{K}_{[c,d)};\mathbb{F})$ is nonzero but the image is in the kernel of $\varphi_{d}: H_{i}(\mathcal{K}_{[c,d)};\mathbb{F}) \to H_{i}(\mathcal{K}_{[d,e)};\mathbb{F})$ for the successor $[d,e) \in I$. Hence, we may assign the union of the interval $[a,d)=\bigcup_{[e,f) \in I, a<e<c}$ to $\gamma$. The $i$-th \emph{persistent diagram} $D(PH_{i}(\ptcloudx ; \mathbb{F}))$ of the persistent homology is a set of all $(a,d) \in \R_{\geq 0}^{2}$ assigned to each homology class. For the economy of notation, let $D_{i}(d_{\ptcloudx}):=D(PH_{i}(\ptcloudx ; \mathbb{F}))$.

Lastly, one can measure the difference between two persistent diagrams from a single point cloud as follows; take a bijection between the two and sum all the distances between a point in the domain and its corresponding image. Given $1 \leq p,q \leq \infty$, the $p$-th \emph{Wasserstein distance} $W_{p}(D_{i}(f),D_{i}(g))$ between two persistent diagrams $D_{i}(f)$ and $D_{i}(g)$ over $q$-norm is the infimum of the differences among all bijections. In other words, $$W_{p}(D_{i}(f),D_{i}(g)) := \underset{\phi \in \Phi(f,g)}{\inf} \left( \sum_{x \in D_{i}(f)}\norm{x-\phi(x)}_{q}^{p}\right)^{1/p}$$
where $\Phi(f,g) = \lbrace \phi | \phi : D_{i}(f) \rightarrow D_{i}(g) \text{ and } \phi \text{  is bijection}\rbrace $, and $\norm{ \cdot}_{q}$ is the usual $q$-norm over a Euclidean space. If $p$ goes to infinity, then the distance is called the \emph{Bottleneck distance}.

\section{\textsc{SPRED} Algorithm}\label{sec:algorithm}

In this section, we propose \textsc{spred} algorithm for shape-preserving linear dimension reduction based on the discrepancy between persistent homologies. The problem objective is to find a linear map $\mathbf{L}$ that best preserves $j$-th homological information from $\mathbb{R}^n$ to $\mathbb{R}^k$ given a high-dimensional point cloud $\mathbb{X} \subset \mathbb{R}^n$. This translates to minimize the $p$-th Wasserstein distance between two persistence diagrams before and after projection. We coin the objective like the following optimization problem,
\begin{equation}\label{eq:opt_basic}
\min_{P \in St(n,k)} f_j (P) := \mathcal{W}_p (D_j (d_\mathbb{X}), D_j(d_\mathbb{Y}))
\end{equation}
where $\mathbb{Y} = \mathbf{L}(\mathbb{X}) = \mathbb{X} P$ is a projected point cloud in low-dimensional space and $St(n,k) = \lbrace Y \in \bbR^{n\times k}|Y^\top Y=I_k \rbrace$ is a Stiefel manifold, the set of all orthonormal $k$-frames in $\mathbb{R}^n$.

\subsection{Basic Algorithm}
From a computational perspective, it is not an attractive option to adopt algorithms that require gradient or Hessian information to solve \eqref{eq:opt_basic} since it is not differentiable. One remedy would be to approximate a Euclidean gradient $\nabla f_j (P^{(t)})$ via the finite difference scheme. However, this exponentiates the computational cost by order of $nk$ while at each iteration, evaluating the cost function involves the construction of persistent diagrams and computing Wasserstein distance, two of which are both expensive operations. 

To alleviate such issues, we use a Riemannian variant of the simulated annealing algorithm \citep{kirkpatrick_optimization_1983-1}. Simulated annealing (SA) is one of the most popular derivative-free algorithms to find the global optimum of a given function whose convergence properties have been well studied \citep{faigle_note_1991, locatelli_convergence_2000, yang_convergence_2000}. A generic version of the SA algorithm for minimization of a real-valued function on the Stiefel manifold is presented in Algorithm \ref{code:algorithm_SA} with a standard choice of the exponential cooling for the temperature schedule.

\begin{algorithm}[ht]
	\caption{Simulated Annealing on the Stiefel Manifold}
	\label{code:algorithm_SA}
	\begin{algorithmic}
		\REQUIRE cost function $f:St(n,k) \rightarrow \mathbb{R}$, temperature $\tau$, cooling parameter $\gamma$, random-walk size $\epsilon$
		\ENSURE a final output $P \in St(n,k)$
		\STATE Construct an initial solution $P$ 
		\WHILE {$\tau > \text{terminal temperature}$}
		\STATE $P^* = \textsf{PERTURB}(P, \epsilon)$
		\STATE $\delta f = f(P^*) - f(P)$
		\IF {$\delta f < 0$}
		\STATE $P \leftarrow P^*$
		\ELSE 
		\STATE Draw a random number $u \sim Uniform(0,1)$
		\IF {$u < \exp \left(-\delta f / \tau\right)$}
		\STATE $P \leftarrow P^*$
		\ENDIF
		\ENDIF
		\STATE $\tau \leftarrow \gamma \cdot \tau$
		\ENDWHILE
	\end{algorithmic}
\end{algorithm}

A key element in adapting the SA algorithm to the Stiefel manifold is neighbor search denoted named as \textsf{PERTURB} procedure in Algorithm \ref{code:algorithm_SA}. A natural choice of random walks on a Riemannian manifold $\mathcal{M}$ is to draw a vector on the tangent space at an iterate $P$ and project it onto the manifold via exponential map $\exp_p : T_p \mathcal{M} \rightarrow \mathcal{M}$. However, the Stiefel manifold does not admit a closed-form formula for an exponential map and relies on expensive numerical schemes \citep{zimmermann_matrix-algebraic_2017-1}. Instead, we take a view of the Stiefel manifold as a submanifold embedded in $\mathbb{R}^{n\times k}$. Numerically speaking, we add $\epsilon_{i,j} \sim \mathcal{N}(0, \sigma^2)$ independently to every entry of the current iterate matrix $P^{(t)}$, apply QR decomposition, and take the orthogonal matrix as a candidate $P^*$.

\subsection{Distributed Algorithm}

One practical issue is that computational cost in evaluating the cost function \eqref{eq:opt_basic} grows exponentially as $\vert \mathbb{X} \vert \rightarrow \infty$, which makes the optimization for an optimal linear projection computationally intractable for a large dataset. Recently, a distributed learning framework was proposed for robust and scalable estimation of a manifold-valued estimator via geometric median  \citep{lin_robust_2020}. First, we divide the data $\mathbb{X}$ into $m$ subsets of roughly equal sizes $\mathbb{X}_1, \ldots, \mathbb{X}_m$ such that $\mathbb{X} = \cup_{i=1}^m \mathbb{X}_i$ and $\mathbb{X}_i \cap \mathbb{X}_j = \emptyset$ for $i\neq j$. Denote $P_i, \ldots, P_m$ be the optimizers of the cost function from each subset $\mathbb{X}_1,\ldots, \mathbb{X}_m$ respectively. The geometric median $P_*$ is the minimizer of the following objective
\begin{equation}\label{eq:opt_median}
P_* = \underset{P \in \mathcal{M}}{\arg\min} \sum_{i=1}^m \rho (P, P_i)
\end{equation}
where $\rho(x, y)$ is a distance for $x,y\in\mathcal{M}$ on a geodesically complete Riemannian manifold. If a collection of estimators is weakly concentrated, the following theorem shows that the geometric median provides a tighter deviation bound.

\begin{theorem}[Theorem 3.1 of \cite{lin_robust_2020}]\label{thm:robust}
	Let $P_1, \ldots, P_m$ be a collection of subset estimators for the parameter $P$ and let the geometric median $P_* = \textrm{med}(P_1,\ldots,P_m)$. Assume a map $\log_{P_*} : B(P_*, \epsilon) \rightarrow T_{P_*}\mathcal{M}$ is $K$-Lipschitz continuous.  If
	\begin{gather}
	P(\rho(P_i, P) > \epsilon) \leq \eta ~\textrm{  for all  }~ i=1,\ldots,m
	\end{gather}
	where $\eta < \alpha \in (0, 1/2)$, then 
	\begin{gather}
	P\left(\rho(P_*, P) > C_\alpha \epsilon \right) \leq \exp (-m \phi (\alpha, \eta))
	\end{gather}
	where 
	\begin{align*}
	C_\alpha &= K (1-\alpha) \sqrt{\frac{1}{1-2\alpha}}\\
	\phi(\alpha,\eta) &= (1-\alpha)\log \frac{1-\alpha}{1-\eta} + \alpha \log \frac{\alpha}{\eta}.
	\end{align*}
\end{theorem}

Given subset estimates $P_1, \ldots, P_m$, we take a geometric median on the Grassmann manifold $Gr(n,k) = \lbrace \text{span}(Y)~|~Y \in \mathbb{R}^{n\times k}, Y^\top Y = I_k \rbrace$ which is a collection of $k$-subspaces. A primary reason for using Grassmann geometry is that numerical operations on the manifold have analytic expressions, unlike those for the Stiefel manifold \citep{edelman_geometry_1998}. The persistence homology induced by a projection matrix is invariant under the right action of the orthogonal group, which justifies the choice of Grassmannian geometry. 

We briefly mention two key operations on a Grassmann manifold \citep{bendokat_grassmann_2020}. First, the exponential map $\text{Exp}_X:\mathcal{T}_X Gr(n,k) \rightarrow Gr(n,k)$ projects a tangent vector $\Delta \in \mathcal{T}_X Gr(n,k)$ onto the manifold itself. Let $\textsf{svd}$ a thin singular value decomposition (SVD) routine and $U\Sigma V^\top = \textsf{svd}(\Delta)$. Then, the exponential map is given by
	\begin{equation}\label{eq:grassmann_exp}
	\text{Exp}_X (\Delta) = \begin{bmatrix}
	XV & U
	\end{bmatrix}
	\begin{bmatrix}
	\cos (\Sigma t)\\
	\sin (\Sigma t)
	\end{bmatrix}
	V^\top
	\end{equation}
	where trigonometric functions $\sin$ and $\cos$ are applied to diagonal entries elementwise. An inverse of the exponential map is called the logarithm map $\text{Log}_X : Gr(n,k) \rightarrow \mathcal{T}_X Gr(n,k)$. For $X,Y \in Gr(n,k)$, the cosine-sine decomposition employs the generalized SVD to compute the pair of SVDs
	\begin{equation*}
	\begin{bmatrix}
	X^\top Y \\
	(I_n - XX^\top ) Y
	\end{bmatrix}
	= \begin{bmatrix}
	V \cos (\Sigma) V^\top\\
	U \sin(\Sigma) V^\top
	\end{bmatrix}
	\end{equation*}  
	where $I_n$ is an $(n\times n)$ identity matrix \citep{edelman_geometry_1998}. Then, we have the logarithmic map by
	\begin{equation}\label{eq:grassmann_log}
	\text{Log}_X (Y) = U\Sigma V^\top.
	\end{equation}

Now we proceed to compute a geometric median using the Weiszfeld algorithm that has been long known and extended to non-Euclidean settings \citep{weiszfeld_sur_1937, eckhardt_webers_1980, fletcher_geometric_2009, aftab_generalized_2015}. The Weiszfeld algorithm may be viewed as a gradient descent that does not need to compute the step size, which eliminates computational bottleneck due to accumulated costs in determining a proper step size at each iteration. 
	\begin{algorithm}[!th]
		\caption{Weiszfeld Algorithm for Geometric Median of Subset Estimates}
		\label{code:algorithm_median}
		\begin{algorithmic}
			\REQUIRE subset estimates $\lbrace P_1,\ldots,P_m\rbrace \subset Gr(n,k)$, stopping criterion 
			\ENSURE a geometric median $P_*$
			\STATE Construct an initial solution $P$ 
			\REPEAT {}
			\STATE Update $P$ by
			\begin{equation*}
			P \leftarrow \text{Exp}_P \left(\frac{\sum_{i=1}^m \text{Log}_P (P_i) / \|\text{Log}_P (P_i)\|}{\sum_{i=1}^m 1/\|\text{Log}_P (P_i)\|}\right)
			\end{equation*}
			where $\text{Exp}_P (\cdot)$ and $\text{Log}_P (\cdot)$ are given by Equations \eqref{eq:grassmann_exp} and \eqref{eq:grassmann_log}
			\UNTIL convergence
		\end{algorithmic}
	\end{algorithm}
	
	The Weiszfeld algorithm for subset estimates on a Grassmann manifold is given in Algorithm \ref{code:algorithm_median}, which has two free components - an initial solution and the convergence assessment. We employ an extrinsic framework via an equivariant embedding that preserves many geometric features on a manifold \citep{bhattacharya_nonparametric_2015-1}. On the Grassmann manifold, the equivariant embedding $\mathcal{J}:Gr(n,k)\rightarrow \mathbb{R}^{n\times n}$ is given by a projection matrix representation, i.e., $\mathcal{J}(P) = PP^\top$. Therefore, we use an extrinsic mean 
	\begin{equation*}
	\bar{P} = \mathcal{J}^{-1} \left( \frac{1}{m} \sum_{i=1}^m \mathcal{J}(P_i)\right) = \mathcal{J}^{-1} \left( \frac{1}{m} \sum_{i=1}^m P_i P_i^\top \right)
	\end{equation*}
	as an initial guess for its ease of computation and convergence of the algorithm is assessed by a small incremental change under the norm $\| \mathcal{P}^{(t)} - \mathcal{P}^{(t+1)} \| < \epsilon$.

\section{Measures of Topological Equivalence}\label{sec:theory}

The use of linear projection $\bfL$ from Equation \eqref{eq:opt_basic} has the advantage that it induces a map between two filtrations $\mathcal{F}_{\ptcloudx}\to \mathcal{F}_{\mathbf{L}(\ptcloudx)}$ for a high-dimensional point cloud $\ptcloudx$. Based on the map, we propose similarity measures to evaluate topological equivalences such as quasi-isomorphism and strong equivalent between two filtrations. Indeed, these similarities can be defined for any filtration homomorphism which will be defined below. 

\begin{definition}[Filtration homomorphism]
	Let $\{\mathcal{K}_{i} \}_{i \in I}$ and $\{\mathcal{Q}_{j} \}_{j \in J}$ be two filtrations of simplicial complexes with totally ordered set $I$ and $J$. Given an order preserving function $f:I \to J$, a \emph{filtration homomorphism} $\phi_{f}:\{\mathcal{K}_{i} \}_{i \in I} \to \{\mathcal{Q}_{j} \}_{j \in J}$ is a set of simplicial maps $\{\phi_{i,j}:\mathcal{K}_{i} \to \mathcal{Q}_{j} \}_{(i,f(i)) \in I \times J}$ such that $\phi_{i',f(i')} \circ (\mathcal{K}_{i} \to \mathcal{K}_{i'}) =(\mathcal{Q}_{f(i)} \to \mathcal{Q}_{f(i')}) \circ \phi_{i,f(i)}$ for any $i,i' \in I$ with $i<i'$.
\end{definition}

Any function $L:\ptcloudx \to \mathbb{Y}$ preserving pairwise distances induces a filtration homomorphism $L^{\ast}:\mathcal{F}_{\ptcloudx}\to \mathcal{F}_{\mathbb{Y}}$. Without loss of generality, we may assume that an order-preserving function $f:I \to J$ in the definition is injective by refinement of $J$. In the case of the linear projection, however, we need a lemma below to derive a canonical filtration homomorphism $\mathcal{F}_{\ptcloudx}\to \mathcal{F}_{\mathbf{L}(\ptcloudx)}$.

\begin{lemma}\label{main1}
	Let $B_{\epsilon}V$ be a union of $\epsilon$-ball over a points in the subspace $V$ in $\R^{n}$. If $\ptcloudx \subseteq B_{\epsilon}V$, then for any two points $\veca,\vecb \in \ptcloudx$, 
	$$\norm{\veca-\vecb}^{2}-4\epsilon^{2} \leq \norm{\matL(\veca)-\matL(\vecb)}^{2} \leq \norm{\veca-\vecb}^{2}.$$
\end{lemma}
\begin{proof}
	Let $\matL$ be a linear projection from $\R^{n}$ to $V \cong \R^{k}$ with $n>k$. We may regard that $V$ lies in $\R^{n}$ by suitable basis change. Moreover, by the equivalence of all norms over finite-dimensional vector spaces, we may assume that $\im\matL$ is spanned by the first $k$ elements of the standard basis up to change of basis. Thus we may assume $\matL$ is a linear transformation sending $\vecv:=(x_{1},x_{2},\cdots, x_{n})$ to $\matL(\vecv)=(x_{1},\cdots, x_{k},0,\cdots,0)$ and define $\boldsymbol{\epsilon}_{\vecv}:= \vecv-\matL(\vecv)$ for any $\mathbf{v} \in \bbR^n$. Then, for any given points $\veca = (x_{1},\cdots, x_{n})$ and $\vecb = (y_{1},\cdots, y_{n})$ in $\ptcloudx$,
	\begin{align*}
	\norm{L(\veca)-L(\vecb)}^{2} &= \sum_{i=1}^{k}(x_{i}-y_{i})^{2} = \norm{\veca-\vecb}^{2}- \norm{\boldsymbol{\epsilon}_{\veca}-\boldsymbol{\epsilon}_{\vecb}}^{2}
	\\
	&=\norm{\veca-\vecb}^{2} -\norm{\boldsymbol{\epsilon}_{\veca}}^{2}-\norm{\boldsymbol{\epsilon}_{\veca}}^{2} + 2\norm{\boldsymbol{\epsilon}_{\veca}}\norm{\boldsymbol{\epsilon}_{\vecb}}\cos(\theta).
	\end{align*}
	For any $\veca,\vecb \in B_{\epsilon}V$, we have the inequality  $\norm{\boldsymbol{\epsilon}_{\veca}}, \norm{\boldsymbol{\epsilon}_{\vecb}} < \epsilon$, which completes the proof.
\end{proof}

Lemma \ref{main1} implies that the inequality $0 \leq \eta_{\min} \leq \eta_{\max} < 4\epsilon^{2}$ holds where
\begin{align*}
\eta_{\max} &= \max_{(\veca,\vecb) \in \ptcloudx^{2}}\norm{\veca-\vecb}-\norm{\matL(\veca)-\matL(\vecb)}\\
\eta_{\min} &= \min_{(\veca,\vecb) \in \ptcloudx^{2}}\norm{\veca-\vecb}-\norm{\matL(\veca)-\matL(\vecb)}.
\end{align*}
Also, assume that $\mathcal{F}_{\ptcloudx}=\{ \mathcal{K}_{[a,b)}\}_{[a,b) \in I}$ and $\mathcal{F}_{\mathbf{L}(\ptcloudx)}=\{ \mathcal{Q}_{[a',b')}\}_{[a',b') \in J}$. For $t \in \R_{\geq0}$, let $\mathcal{K}_{t}$ (or $\mathcal{Q}_{t}$) be $\mathcal{K}_{[a,b)}$  (or $\mathcal{Q}_{[a',b')}$) such that $t \in [a,b) \in I$ (or $t \in [a',b') \in J$) respectively. Moreover, $\sigma:=\{\veca_{0},\cdots, \veca_{m} \} \in \mathcal{K}_{t}$ with $t \in [a,b) \in I$ implies $\matL(\sigma):=\{\matL(\veca_{0}),\cdots, \matL(\veca_{m}) \} \in \mathcal{Q}_{t'}$ for some $t'>t-\eta_{\min}$. Hence, $\matL$ induces a filtration homomorphism as below.

\begin{definition}[Canonical embedding]\label{canon_emb}
	Given a linear projection $\matL$ and real number $0 \leq \eta<\eta_{\min}$, the \emph{canonical embedding} $\matL^{\ast}_{\eta}:\mathcal{F}_{\ptcloudx}\to \mathcal{F}_{\mathbf{L}(\ptcloudx)}$ is defined by a simplicial map $\sigma \in \mathcal{K}_{t+\eta} \mapsto \matL(\sigma) \in \mathcal{Q}_{t}$ for each $t \in \R_{\geq0}$. By replacing $I$ and $J$ with a suitable refinement of $I \cup J$ represented as $I=\{[a_{0},a_{1}),[a_{1},a_{2}), \cdots, [a_{n-1}, \infty) \}$ and $J= \{[a_{0}-\eta,a_{1}-\eta),[a_{1}-\eta,a_{2}-\eta), \cdots, [a_{n-1}-\eta, \infty) \}$ for some $n \in \N$, we may write $\matL^{\ast}_{\eta}$ as a set of simplicial maps
	$$\matL^{\ast}_{\eta}:=\{ \left(\matL^{\ast}_{\eta}\right)_{i}:\mathcal{K}_{[a_{i},a_{i+1}) } \to \mathcal{Q}_{[a_{i}-\eta,a_{i+1}-\eta)}\}_{ i \in \{0,1, \cdots, n\} }$$ 
	for some $n \in \N$.
\end{definition}
We note that $\matL^{\ast}_{\eta}$ is still a finite set. The relation above gives an apparatus to compare filtrations by Theorem \ref{classification_thm}. For notational simplicity, let $\mathcal{K}_{i}$ and $\mathcal{Q}_{i}$ denote $\mathcal{K}_{[a_{i},a_{i+1}) }$ and $\mathcal{Q}_{[a_{i}-\eta,a_{i+1}-\eta)}$, respectively. Denote $\mathcal{K}^{l}$ for the $l$-skeleton of any given simplicial complex $\mathcal{K}$.
\begin{theorem}\label{classification_thm}
	Given $\left(\left(\matL^{\ast}_{\eta}\right)_{i}:\mathcal{K}_{i} \to \mathcal{Q}_{i}\right) \in \matL^{\ast}_{\eta}$ and $l \in \N \cup \{ \infty\}$ the followings are equivalent.
	\begin{enumerate}
		\item[(1)] $\left(\matL^{\ast}_{\eta}\right)_{i}$ induces $H_{j}(\mathcal{Q}_{i}^{l}) \cong H_{j}(\mathcal{K}_{i}^{l})$ for all $j \leq l$.
		\item[(2)] $H_{j}(\mathcal{Q}_{i}^{l}/\im\left(\matL^{\ast}_{\eta}\right)_{i})=0$ for all $j \leq l$.
	\end{enumerate}
	If the fundamental group $\pi_{1}(\mathcal{Q}_{i}^{l}/\im\left(\matL^{\ast}_{\eta}\right)_{i},x)$ with respect to every $x \in\mathcal{Q}_{i}^{l}/\im\left(\matL^{\ast}_{\eta}\right)_{i}$ is trivial, then
	\begin{enumerate}[resume]
		\item[(3)] $\mathcal{K}_{i}^{l}$ and $\mathcal{Q}_{i}^{l}$ are homotopic equivalent.
	\end{enumerate}
	is equivalent to (1) and (2).
\end{theorem}
\begin{proof}
	It suffices to show the case $l=\infty$ by replacing original filtrations with filtrations consisting of $l$-skeletons.  To see the equivalence of (1) and (2), Proposition A.5 of \cite{hatcher} asserts that $\left(\im \left(\matL^{\ast}_{\eta}\right)_{i}, \mathcal{Q}_{i}\right)$ is a good pair in that we have $H_{j}\left(\mathcal{Q}_{i}/\im\left(\matL^{\ast}_{\eta}\right)_{i}\right) \cong H_{j}\left(\mathcal{Q}_{i}, \im\left(\matL^{\ast}_{\eta}\right)_{i}\right)$ as a natural consequence \cite[Proposition 2.22]{hatcher}. Therefore, the acyclicity of $\mathcal{Q}_{i}/\im\left(\matL^{\ast}_{\eta}\right)_{i}$ holds if and only if $H_{k}(\mathcal{Q}_{i}) \cong H_{j}\left(\im\left(\matL^{\ast}_{\eta}\right)_{i}\right)$ by the long exact sequence of relative simplicial homology. Finally, $\matL^{\ast}_{\eta}(\mathcal{K}_{i})$ and $\mathcal{K}_{i}$ are homotopically equivalent because $\left(\matL^{\ast}_{\eta}\right)_{i}$ is a simplicial map.
	
	The direction from (3) to (2) is trivial. From (2) to (3) is the consequence of the proof of the Hurewicz theorem stated in \cite[Corollary 4.33]{hatcher} applied for each path-connected component. 
\end{proof}

By Theorem \ref{classification_thm}, for each $\left(\left(\matL^{\ast}_{\eta}\right)_{i}:\mathcal{K}_{i} \to \mathcal{Q}_{i}\right) \in \matL^{\ast}_{\eta}$ for $0 \leq j \leq l$ with fixed $l \in\N$, only one of the following statements holds,
\begin{enumerate}
	\item $H_{j}(\mathcal{Q}_{i}^{l}) \not\cong H_{j}(\mathcal{K}_{i}^{l})$
	\item $H_{j}(\mathcal{Q}_{i}^{l}) \cong H_{j}(\mathcal{K}_{i}^{l})$ \text{ and } $\exists x \in \mathcal{Q}_{i}^{l}/\im\left(\matL^{\ast}_{\eta}\right)_{i}$ such that $\pi_{1}(\mathcal{Q}_{i}^{l}/\im\left(\matL^{\ast}_{\eta}\right)_{i},x) \neq0$
	\item $H_{j}(\mathcal{Q}_{i}^{l}) \cong H_{j}(\mathcal{K}_{i}^{l})$ \text{ and } $\pi_{1}(\mathcal{Q}_{i}^{l}/\im\left(\matL^{\ast}_{\eta}\right)_{i},x) =0$ for any $x \in \mathcal{Q}_{i}^{l}/\im\left(\matL^{\ast}_{\eta}\right)_{i}$.
\end{enumerate}

Therefore, we have two similarities measuring topological differences between two filtrations based on a partition of $\R_{\geq 0}$. For notational efficiency, we write down $\pi_{1}(\mathcal{Q}_{i}^{l}/\im\left(\matL^{\ast}_{\eta}\right)_{i}) \neq 0$ if there exists $x \in \mathcal{Q}_{i}^{l}/\im\left(\matL^{\ast}_{\eta}\right)_{i}$ such that $\pi_{1}(\mathcal{Q}_{i}^{l}/\im\left(\matL^{\ast}_{\eta}\right)_{i},x) \neq0$. 
\begin{definition}[Similarities over the topological differences]
	Given $\matL^{\ast}_{\eta}$ as in Definition \ref{canon_emb} and $l \in \N$, define a partition of  $\R_{\geq 0}$ by
	\begin{align*}
	T_{0}&:=  \bigcup_{\substack{i \in \{1,2, \cdots, n-1\} \\H_{j}(\mathcal{Q}_{i}^{l}) \not\cong H_{j}(\mathcal{K}_{i}^{l}) } }[a_{i},a_{i+1}), & T_{1}&:= \bigcup_{\substack{i \in \{1,2, \cdots, n-1\} \\H_{j}(\mathcal{Q}_{i}^{l}) \cong H_{j}(\mathcal{K}_{i}^{l}) \\ \pi_{1}(\mathcal{Q}_{i}^{l}/\im\left(\matL^{\ast}_{\eta}\right)_{i}) \neq 0 } }[a_{i},a_{i+1}),  &
	T_{2}&:=\bigcup_{\substack{i \in \{1,2, \cdots, n-1\} \\H_{j}(\mathcal{Q}_{i}^{l}) \cong H_{j}(\mathcal{K}_{i}^{l}) \\ \pi_{1}(\mathcal{Q}_{i}^{l}/\im\left(\matL^{\ast}_{\eta}\right)_{i}) =0} }[a_{i},a_{i+1}) 
	\end{align*}
	which are all disjoint to each other by Theorem \ref{classification_thm} and $\bigcup_{i=0}^{2}T_{i}=[0,a_{n})$ where $a_{n}= \diam \ptcloudx /2$. We may define the following quantities
	\begin{align*}
	\mu_{\operatorname{quasi-iso}}(\matL^{\ast}_{\eta}) &:= |T_{1}\cup T_{2}|/a_{n} & \mu_{\operatorname{equiv}}(\matL^{\ast}_{\eta}) &:= |T_{2}|/a_{n}
	\end{align*}
	where $|\cdot|$ is a standard Lebesgue measure of $\R$. We call $\mu_{\operatorname{quasi-iso}}(\matL^{\ast}_{\eta})$ and $\mu_{\operatorname{strong}}(\matL^{\ast}_{\eta})$ as \emph{quasi-isomorphism} similarity and \emph{strong-equivalence} similarity, respectively.
\end{definition}
These similarities measure how much topological properties such as homology classes or homotopic equivalence are preserved during the constructions of Rips or \v{C}ech complexes via the projection.

Lastly, we propose algorithms to calculate each similarity. First, an algorithm for  $\mu_{\operatorname{quasi-iso}}(\matL^{\ast}_{\eta})$ can be derived by comparing the persistent diagrams $D_{k}(d_{\ptcloudx})$ and $D_{k}(d_{\ptcloudy})$ or barcodes of them as below.

\begin{proposition}
	Given $\eta$, there is an algorithm to calculate $\mu_{\operatorname{quasi-iso}}(\matL^{\ast}_{\eta})$ using the persistent diagram $D_{k}(d_{\ptcloudx})$ and $D_{k}(d_{\ptcloudy})$ or barcodes of $\ptcloudx$ and $\ptcloudy$ for $k \leq l$.
\end{proposition}
\begin{proof}
	Suppose we are given the persistent diagrams. Fix $k$. Let $A_{d_{\ptcloudx}}:=\{a \in \R : (a,b) \text{ or } (b,a) \in  D_{k}(d_{\ptcloudx}) \text{ for some }b \in \R \}.$ In other words, $A_{d_{\ptcloudx}}$ is a set of all real numbers appearing in the tuple of $D_{k}(d_{\ptcloudx})$. Let $A=\{ a-\eta: a \in A_{d_{\ptcloudx}} \}\cup A_{d_{\ptcloudy}}$. By ordering $A$ we may assume $A=\{ -\eta=a_{0} < a_{1}  < \cdots < a_{n}=t \}$ where $t:= \min(\diam\ptcloudx,\diam\ptcloudy)/2$. For each $i \in \{ 0,1, \cdots, n-1\}$, let $P_{i,\ptcloudx}:=\{(a,b) \in D_{k}(d_{\ptcloudx}): a\leq a_{i}+\eta< b \}$ and $P_{i,\ptcloudy}:=\{(a,b) \in D_{k}(d_{\ptcloudx}): a\leq a_{i}< b \}$. If $|P_{i,\ptcloudx}| = |P_{i,\ptcloudy}|$, then $H_{i}(\mathcal{K}_{[a_{i}-\eta,a_{i+1}-\eta)}) \cong H_{i}(\mathcal{Q}_{[a_{i},a_{i+1})})$. Therefore, $$\mu_{\operatorname{quasi-iso}}(\matL^{\ast}_{\eta}) \cdot \diam \ptcloudx/2 = \sum_{\substack{i \in \{ 0,1, \cdots, n-1\} \\ |P_{i,\ptcloudx}| = |P_{i,\ptcloudy}| }} (a_{i+1}-a_{i}).$$
	
If one has barcodes $B_{\ptcloudx}$ and $B_{\ptcloudy}$ of $\ptcloudx$ and $\ptcloudy$ respectively, we may define \emph{height} of barcode $\httt_{t} B_{\ptcloudx}$  at time $t$ as the number of bars appearing at $t$. Then, $$\mu_{\operatorname{quasi-iso}}(\matL^{\ast}_{\eta}) \cdot \diam \ptcloudx/2= \norm{\{t \in [0,\diam \ptcloudx/2): \httt_{t-\eta} B_{\ptcloudx} =\httt_{t} B_{\ptcloudy}\}}$$ 
which can be calculated by the similar procedure above, since $B_{\ptcloudx}$ and $B_{\ptcloudy}$ have finitely many distinct heights.
\end{proof}

Next, we calculate $\pi_{1}(\mathcal{Q}_{i}^{l}/\im\left(\matL^{\ast}_{\eta}\right)_{i},x)$ as a finitely presented group. This uses an algorithm adopted by \texttt{SageMath}'s Finite Simplicial complexes and Finitely presented groups package \citep{sagemath}. 
\begin{proposition}
	There is an algorithm computing $\pi_{1}(\mathcal{Q}_{i}^{l}/\im\left(\matL^{\ast}_{\eta}\right)_{i},x)$ as a finitely presented group.
\end{proposition}
\begin{proof}
	For a connected simplicial complex $\mathcal{K}$, the edge-path group of $\mathcal{K}$ is isomorphic to $\pi_{1}(\mathcal{K})$ \citep{MR1954333,sagemath}. The edge-path group can be finitely presented using Kruskal's algorithm and found by Finite Simplicial complexes package of \texttt{SageMath}. Hence, for each path-connected component $Q$ of $\mathcal{Q}_{i}^{l}$ containing $x$, we can find a path-connected component $K$ of $\mathcal{K}_{i}^{l}$ mapped into $Q$. This gives us the exact sequence 
	$$\pi_{1}(K) \to \pi_{1}(Q) \to \pi_{1}(\mathcal{Q}_{i}^{l}/\im\left(\matL^{\ast}_{\eta}\right)_{i},x) \to \pi_{0}(K)=0$$ 
	from the long exact sequence of relative fundamental groups. Therefore, $\pi_{1}(\mathcal{Q}_{i}^{l}/\im\left(\matL^{\ast}_{\eta}\right)_{i},x)$ is the cokernel of the map $\pi_{1}(K) \to \pi_{1}(Q)$, which can be calculated in \texttt{SageMath}.
\end{proof}

We close this section with a remark. It is generally not decidable to figure out whether the finitely generated group is trivial or not \citep{MR1307623}[Theorem 12.32]. However, we are certain that such a case does rarely happens since this projection map usually gives the maximal-dimensional simplex only.

\section{Examples}\label{sec:example}

In this section, we consider two examples. The first example simulates the data from a cylinder $\mathbb{S}^1 \times [-2,2]$ with additive Gaussian noise with zero mean and variance 0.05 \citep{kachan_persistent_2020}. A cylinder in $\mathbb{R}^3$ has Betti numbers $\beta_0 = 1$ and $\beta_1 = 1$. This means that the \textsc{spred} algorithm of orders 0 and 1 should be able to capture a connected component and a single hole, respectively. We run \textsc{spred} algorithm of orders 0 and 1 on the generated data and compare it with PCA. 

\begin{figure}[ht]
	\centering
	\includegraphics[width=0.95\linewidth]{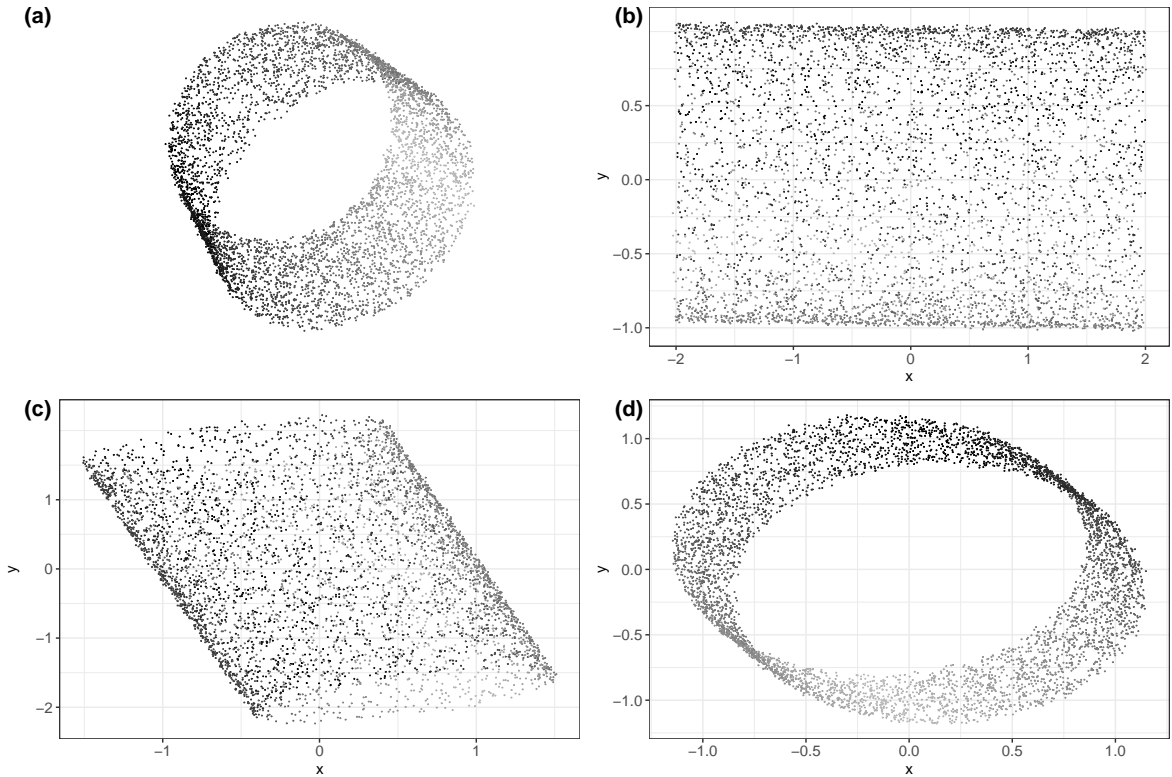}
	\caption{Example of a noisy cylinder $\mathbb{S}^1 \times [-2,2]$ with additive noise $\mathcal{N}(0,0.01)$. (a) The data is shown in $\mathbb{R}^3$ and projected data are shown with (b) PCA, and \textsc{spred} algorithms of (c) order 0 and (d) order 1.}
	\label{fig:cylinder}
\end{figure}
The results are visualized in Figure \ref{fig:cylinder}, where PCA shows a rectangular shape of the projected point cloud. This corresponds to the third axis of the data generating process $[-2,2]$ which has the largest degree of variation. On the other hand, \textsc{spred} algorithm of $j=0$ captures the connected pattern at a slanted perspective. An interesting observation is attained through \textsc{spred} algorithm of order 1 where an interior of the cylinder is clearly shown as an ellipse in its 2-dimensional projection.

\begin{figure}[ht]
	\centering
	\includegraphics[width=0.95\linewidth]{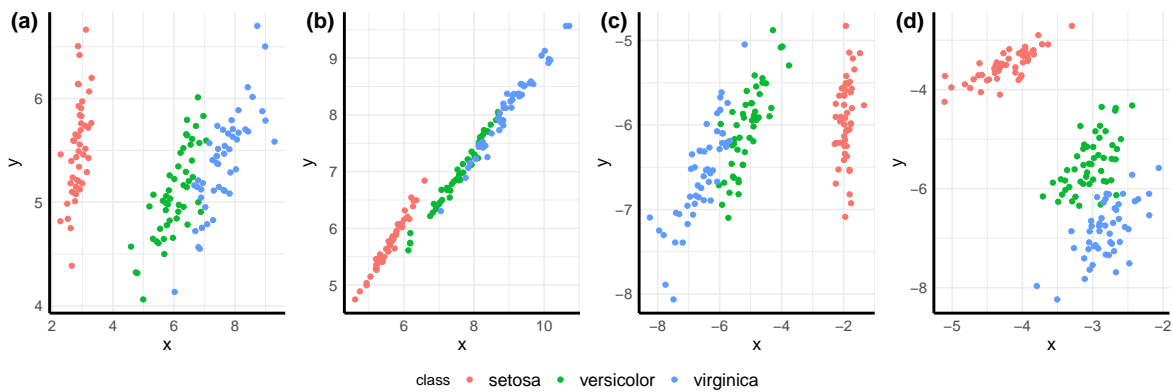}
	\caption{Example of \textit{iris} dataset. Projections onto $\bbR^2$ are shown for (a) PCA, (b) random projection, and \textsc{spred} algorithm of (c) order 0 and (d) order 1.}
	\label{fig:iris}
\end{figure}

Next, we took an example of the famous \textit{iris} dataset \citep{anderson_species_1936, fisher_use_1936-1}, which gives the measurements of four morphological variables for 50 flowers from each of the 3 species of iris. We compare \textsc{spred} algorithm with PCA and random projection \citep{beals_extensions_1984-1}, which are shown in Figure \ref{fig:iris}.

It is known that the iris dataset has a natural division of the species to separate Setosa class and the other two. All four algorithms show a similar pattern to make the consistent separation while \textsc{spred} algorithm has two distinct features. First, \textsc{spred} of order 0 penalizes dispersion of largely deviated observations so that it focuses on a compact representation of connected components. The 1st-order \textsc{spred} algorithm seems to make a more clear distinction between Versicolor and Virginica classes as two classes are more circular-shaped with a smaller relative area of the intersection than others.

\section{Conclusion}\label{sec:conclusion}

We proposed a topology-based dimension reduction framework that explicitly recovers a linear projection preserving persistent homology between the original data in high-dimensional space and its embedding in low-dimensional space. This perspective motivated a systematic approach to compare simplicial complexes by a filtration homomorphism that led to topological similarity measures $\mu_{\operatorname{quasi-iso}}$ and $\mu_{\operatorname{equiv}}$. Based on a gradient-free optimization routine, \textsc{spred} algorithm showed successful results with respect to the preservation of topological features on two examples.

One notable benefit of our framework is that a gradient-free formulation does not require tailored derivation in merging heterogeneous information into the linear dimension reduction framework. For example, balancing the $0$-th and $1$-st order persistent homology can be achieved by setting a cost function $f_{\lambda}(P) = \lambda \cdot \mathcal{W}_p (D_0 (d_\bbX), D_0 (d_\mathbb{Y}) + (1-\lambda) \cdot \mathcal{W}_p (D_1 (d_\bbX), D_1 (d_\mathbb{Y}))$ for some $\lambda \in (0,1)$ that controls how much information should be weighed for each order. This formulation can be further extended to reflect other types of information that have been objectives of many linear dimensionality reduction algorithms in the forms of minimization problems over the Stiefel manifold \citep{cunningham_linear_2015}. For instance, one may regularize \textsc{spred} algorithm by adding a penalty term $\text{tr}(P^\top \Sigma_\bbX P)$ where $\Sigma_{\bbX}$ is an empirical covariance matrix in the spirit of PCA.

Devising an efficient computational pipeline is an interesting direction for future studies. The \textsc{spred} algorithm consists of several costly modules such as the reconstruction of simplicial complexes, measuring the difference of two via Wasserstein distance, and stochastic optimization over a Stiefel manifold. When the data contains a higher level of noise and is of large volume, the first two components become exponentially lavish and persistence diagrams become large and arouse high costs even with fast-evolving tools from computational optimal transport for a single evaluation. One remedy is to reduce the scale of computation at each component. For instance, the 2-Wasserstein distance between two persistence diagrams may be approximated by a two-step approach to represent each diagram's point set as a mixture of Gaussian distributions and compute the distance as if they are weighted point sets on the manifold of Gaussian distributions \citep{chen_optimal_2019-1}. Another subject of future research is to exploit an efficient stochastic optimization on a Stiefel manifold. Moreover, implementing an algorithm calculating $\mu_{\operatorname{equiv}}$ using available mathematical software such as \texttt{SageMath} \citep{sagemath} and \texttt{Perseus} from \cite{perseus} needs to be done in the future.

\bibliographystyle{dcu}
\bibliography{PHdimred}

\end{document}